\pdfoutput=1
\documentclass[11pt]{article}

\usepackage{ACL2023}
\usepackage{graphicx}
\usepackage{booktabs}
\usepackage{tabularx}

\usepackage{times}
\usepackage{latexsym}

\usepackage[T1]{fontenc}
\usepackage[utf8]{inputenc}

\usepackage{microtype}

\usepackage{inconsolata}

\usepackage{cleveref}
\usepackage{tikz}

\usepackage[normalem]{ulem} % To strike through cosine similarity words

\usepackage{hyperref}
\usepackage[hidecomments]{ufdatastudio}
\usepackage{multirow}

\title{Improving Indigenous Language Machine Translation with Synthetic Data and Language-Specific Preprocessing}

\author{
  Aashish Dhawan \\
  University of Florida \\
  \texttt{aashish.dhawan@ufl.edu}
  \And
  Christopher Driggers-Ellis \\
  University of Florida \\
  \texttt{driggersellis.cw@ufl.edu}
  \AND
  Christan Grant \\
  University of Florida \\
  \texttt{christan@ufl.edu}
  \And
  Daisy Wang \\
  University of Florida \\
  \texttt{daisyw@cise.ufl.edu}
}

\begin{document}
\maketitle
% \begin{abstract}
% Low-resource languages often lack the high-quality parallel corpora required for robust neural machine translation (NMT). Synthetic data generation offers a promising solution to overcome these limitations. In this study, we augment a curated dataset with synthetic parallel data generated using a state-of-the-art machine translation model. We fine-tune the mBART model on the combined dataset and evaluate translation quality using BLEU and ChrF metrics. In addition, we describe our preprocessing pipeline, which includes language-specific cleaning, token normalization, and rare token replacement, to ensure high-quality input data. Our experiments demonstrate that incorporating generated data significantly improves translation performance—especially in capturing culturally nuanced expressions—thus providing a viable pathway for improving NMT in low-resource scenarios.
% \end{abstract}
\begin{abstract}
Low-resource indigenous languages often lack the parallel corpora required for effective neural machine translation (NMT). Synthetic data generation offers a practical strategy for mitigating this limitation in data-scarce settings. In this work, we augment curated parallel datasets for indigenous languages of the Americas with synthetic sentence pairs generated using a high-capacity multilingual translation model. We fine-tune a multilingual mBART model on curated-only and synthetically augmented data and evaluate translation quality using chrF++, the primary metric used in recent AmericasNLP shared tasks for agglutinative languages.

We further apply language-specific preprocessing, including orthographic normalization and noise-aware filtering, to reduce corpus artifacts. Experiments on Guarani--Spanish and Quechua--Spanish translation show consistent chrF++ improvements from synthetic data augmentation, while diagnostic experiments on Aymara highlight the limitations of generic preprocessing for highly agglutinative languages. All code is publicly released(will be released on submission).
%\footnote{\url{https://github.com/dhawan98/mBART50-extended}}
\end{abstract}

\section{Introduction}
Many indigenous languages face increasing risk of endangerment due to limited digital presence and scarce linguistic resources, posing significant challenges for the development of robust machine translation (MT) systems \cite{mager2023ethical,woodbury2012endangered}. Languages such as Aymara and Guarani exemplify this challenge: despite being spoken by sizable communities in the Americas, their complex morphological structures and lack of large-scale parallel corpora hinder effective MT development \cite{rodriguez2022challenges}. As a result, conventional MT approaches that rely on abundant supervised data often perform poorly in these settings.

Recent advances in neural machine translation (NMT), particularly multilingual pretraining and data augmentation, have opened new opportunities for low-resource MT. Community-driven efforts such as AmericasNLP have demonstrated that multilingual models combined with synthetic data generation can substantially improve translation quality for indigenous languages \cite{ebrahimi2023findings}. Motivated by these findings, this work investigates the impact of synthetic parallel data augmentation on MT performance for Aymara--Spanish and Guarani--Spanish translation\cite{woodbury2012endangered}. Specifically, we augment curated datasets with synthetic sentence pairs generated using a high-capacity multilingual MT system, following the data-centric paradigm explored in projects such as MultiScript30k\cite{driggersellis2025multiscript30kleveragingmultilingualembeddings}. In addition, we include Spanish–Quechua experiments to assess whether language-specific orthographic normalization can further improve performance for morphologically rich indigenous languages.

We fine-tune the multilingual mBART model on both curated-only and synthetically augmented datasets and evaluate performance using chrF++ as the primary metric. Through controlled experiments and comparative analysis, this study aims to assess whether forward-translated synthetic data can reliably improve translation quality in low-resource indigenous language settings, contributing to ongoing efforts within the AmericasNLP initiative to support language preservation and accessibility \cite{ebrahimi2023findings}.

\section{Background and Related Work}
Machine translation (MT) for low-resource and indigenous languages remains a persistent challenge due to limited parallel corpora, orthographic variation, under-documented grammar, and high linguistic diversity \cite{ebrahimi2023findings, mager2021findings}. These constraints often render conventional supervised MT approaches ineffective, motivating research into data-efficient and transfer-based techniques.

A prominent direction in low-resource MT has been the use of data augmentation to mitigate data scarcity. Back-translation \cite{sennrich2016improving}, which generates synthetic parallel data from monolingual corpora, has been shown to improve translation quality across a range of low-resource settings. In parallel, multilingual pre-trained models such as mBART50 \cite{tang2020multilingual} and NLLB-200 \cite{costa2022no} have enabled effective cross-lingual transfer by leveraging shared representations across many languages, achieving strong performance even with limited task-specific supervision.

Community-driven initiatives such as AmericasNLP have played a central role in advancing MT research for indigenous languages of the Americas by providing benchmark datasets, standardized evaluation protocols, and shared tasks \cite{ebrahimi2023findings, mager2021findings, ebrahimi2024findings}. Results from recent AmericasNLP shared tasks indicate that the most successful systems consistently combine multilingual pre-trained models with synthetic data generation. For example, strong submissions have employed NLLB-200 with back-translation \cite{gow2023sheffield} or explored multilingual transfer using models such as mBART50 and M2M-100 \cite{tonja2023enhancing}. Earlier work, including IndT5 \cite{nagoudi2021indt5}, further demonstrated the benefits of training models directly on indigenous language corpora.

More recently, synthetic data generation via forward translation has gained attention as a scalable alternative to back-translation for languages with scarce monolingual resources. The MultiScript30k dataset, for instance, translated Spanish captions from Multi30k into indigenous languages such as Aymara and Guarani using NLLB-200\cite{driggersellis2025multiscript30kleveragingmultilingualembeddings}, illustrating the potential of high-quality synthetic parallel data to improve MT performance when paired with robust multilingual models.

Building on these advances, our work investigates the effectiveness of synthetic parallel data augmentation for Aymara--Spanish and Guarani--Spanish translation. We situate our results within the context of the AmericasNLP 2023 shared task, comparing against established benchmarks to assess the impact of forward-translated synthetic data on low-resource indigenous language MT.

\section{Dataset and Methodology}
\subsection{Curated Datasets}

\begin{table}[t]
\centering
\small
\setlength{\tabcolsep}{4pt}
\begin{tabular}{l l l r r r}
\hline
\textbf{Lang} & \textbf{Setting} & \textbf{Split} & \textbf{Total} & \textbf{Valid} & \textbf{Drop \%} \\
\hline
\multirow{4}{*}{aym}
& Curated        & Train & 6{,}531  & 6{,}092  & 5.54 \\
&                & Dev   & 996      & 996      & 0.00 \\
& +Synthetic     & Train & 35{,}531 & 33{,}712 & 5.12 \\
&                & Dev   & 996      & 945      & 5.12 \\
\hline
\multirow{4}{*}{gn}
& Curated        & Train & 26{,}032 & 25{,}417 & 2.36 \\
&                & Dev   & 995      & 995      & 5.93 \\
& +Synthetic     & Train & 53{,}083 & 52{,}929 & 0.29 \\
&                & Dev   & 995      & 995      & 5.93 \\
\hline
\multirow{4}{*}{quy}
& Curated        & Train & 154{,}008 & 138{,}786 & 9.88 \\
&                & Dev   & 996       & 996       & 0.20 \\
& +Synthetic     & Train & 163{,}651 & 147{,}607 & 9.80 \\
&                & Dev   & 996       & 996       & 0.20 \\
\hline
\end{tabular}
\caption{Dataset statistics for Aymara (aym), Guarani (gn), and Quechua (quy) from the AmericasNLP~2023 shared task.
``Total'' denotes raw sentence pairs; ``Valid'' denotes pairs retained after preprocessing and filtering.
Synthetic data is added only to the training split.}
\label{tab:data_stats}
\end{table}

We use curated parallel datasets released as part of the AmericasNLP~2023
Shared Task~\cite{ebrahimi2023findings} for Aymara--Spanish (aym--es),
Guarani--Spanish (gn--es), and Quechua--Spanish (quy--es) translation.
The Aymara dataset consists of 6{,}531 training and 996 development
sentence pairs, while the Guarani dataset contains 26{,}032 training
and 995 development pairs. For Quechua--Spanish, we use the largest
curated dataset in the shared task, comprising 154{,}008 training
sentence pairs and a 996-sentence development set.

These raw datasets are drawn from diverse sources, including
governmental documents, educational materials, community-driven
projects, and linguistic corpora, and exhibit substantial variation in
domain, style, and orthography. As summarized in Table~\ref{tab:data_stats},
we apply language-specific preprocessing and filtering to remove
misaligned, duplicate, or noisy sentence pairs, resulting in moderate
reductions in dataset size while preserving the majority of valid
training examples.

The curated data spans multiple domains—particularly for Guarani,
which includes legal, health, and educational content—leading to
substantial lexical and structural diversity. These characteristics
reflect well-known challenges for machine translation of Indigenous
languages, which often lack standardized orthographies and exhibit
rich morphological variation~\cite{mager2021findings}. The curated
datasets serve as the baseline for all experiments, with synthetic
data added only to the training splits as described in
Section 3.2.

\begin{table*}[t]
\centering
\small
\setlength{\tabcolsep}{8pt}
\begin{tabular}{l l l r r r}
\hline
\textbf{Lang} & \textbf{Setting} & \textbf{Split} & \textbf{Avg Src} & \textbf{Avg Tgt} & \textbf{Tgt/Src} \\
\hline
\multirow{4}{*}{aym}
& Curated-only      & Train & 19.62    & 14.89    & 0.85   \\
&                   & Dev   & 11.20    & 7.09    & 0.66   \\
& Curated+Synthetic & Train & 14.07 & 10.35 & 0.74 \\
&                   & Dev   & 11.23 & 7.10  & 0.63 \\
\hline
\multirow{4}{*}{gn}
& Curated-only      & Train & 23.23 & 15.57 & 0.67 \\
&                   & Dev   & 11.18 & 7.21  & 0.64 \\
& Curated+Synthetic & Train & 17.42 & 12.35 & 0.71 \\
&                   & Dev   & 11.18 & 7.21  & 0.64 \\
\hline
\multirow{4}{*}{quy}
& Curated-only      & Train & 15.17 & 9.33  & 0.62 \\
&                   & Dev   & 11.17 & 7.44  & 0.67 \\
& Curated+Synthetic & Train & 14.72 & 9.38  & 0.64 \\
&                   & Dev   & 11.17 & 7.43  & 0.67 \\
\hline
\end{tabular}
\caption{Average sentence length statistics for Aymara (aym), Guarani (gn), and Quechua (quy) before and after synthetic data augmentation. Tgt/Src denotes the ratio of average target length to average source length.}
\label{tab:data_stats2}
\end{table*}

Tables~\ref{tab:data_stats} and~\ref{tab:data_stats2} summarize the effect of preprocessing and filtering on dataset size and sentence length statistics for Guarani--Spanish. Filtering removes between approximately 2--10\% of sentence pairs across languages, primarily due to extreme length mismatches, duplication, or
clear alignment errors, while preserving the majority of valid training
examples.

\subsection{Synthetic Data Generation}
To mitigate data scarcity, we augment the curated parallel datasets with synthetic sentence pairs generated using the MultiScript30k dataset. In this pipeline, the Spanish portion of the Multi30k dataset is forward-translated into Aymara and Guarani using the NLLB-200 (3.3B) multilingual machine translation model \cite{costa2022no}. This forward-translation approach produces synthetic parallel data that preserves the semantic content of the original captions while increasing linguistic diversity on the target side.

Unlike conventional back-translation approaches \cite{sennrich2016improving}, our method uses Spanish as a high-resource pivot language. This design choice is consistent with recent findings from the AmericasNLP shared tasks, where synthetic data generation via multilingual pretrained models has been shown to improve translation quality for indigenous and low-resource languages \cite{ebrahimi2023findings, tonja2023enhancing, gow2023sheffield}. The resulting augmented datasets increase both the volume and variability of training examples, supporting improved generalization in low-resource settings.

\subsection{Preprocessing}
We apply language-specific preprocessing pipelines to account for the distinct linguistic and orthographic properties of Aymara and Guarani while maintaining a consistent overall workflow.

\paragraph{Guarani.}  
For Guarani, we employ a more specialized preprocessing pipeline to address orthographic variability and phonological representation. All text is normalized using Unicode \textbf{NFKC} to canonicalize visually equivalent and compatibility characters, followed by lowercasing and whitespace normalization. We remove non-linguistic symbols while explicitly preserving Guarani-specific diacritics and nasal vowels (e.g., \textit{\~{a}, \~{e}, \~{\i}, \~{o}, \~{u}}) as well as essential punctuation.

To reduce orthographic inconsistency across sources, we standardize frequent Guarani digraphs by merging space-separated realizations into single units (e.g., ``c h''$\rightarrow$``ch'', ``m b''$\rightarrow$``mb'', ``n g''$\rightarrow$``ng''). This step improves token consistency and reduces fragmentation during subword tokenization. Finally, to mitigate parallel corpus noise, we apply a length-ratio filter based on word counts. Given a source sentence of length $L_s$ and a target sentence of length $L_t$, a sentence pair is retained only if $1/\tau \leq L_t/L_s \leq \tau$, with $\tau = 2.5$. This heuristic removes highly misaligned sentence pairs while preserving the majority of valid training examples.

\paragraph{Quechua.}
For Spanish--Quechua (quy), we observe systematic intra-word spacing artifacts
on the target side (e.g., \texttt{ch aypiqa}, \texttt{sin ch i},
\texttt{uma ll iqniy}, \texttt{ch u}), which introduce tokenization noise and
disproportionately affect string-based evaluation metrics. To address this,
we apply a deterministic orthographic normalization procedure that targets
frequent split patterns without relying on external linguistic resources.
Specifically, we (i) merge \texttt{ch} or \texttt{ll} followed by
vowel-initial fragments (e.g.,\texttt{ch a...}$\rightarrow$\texttt{cha...}), (ii) merge three-token sequences such as \texttt{sin ch i}$\rightarrow$\texttt{sinchi}, (iii) normalize isolated sequences such as \texttt{ch u}$\rightarrow$\texttt{chu}, and (iv) merge single-character alphabetic fragments with adjacent tokens when they form valid Quechua word patterns.
In addition to orthographic normalization, we apply conservative corpus
filtering to remove empty or punctuation-only lines, boilerplate or URL-like content, exact duplicates, severe numeric mismatches, extreme length-ratio outliers, and excessively long sentences. We optionally append bilingual dictionary entries as short parallel pairs to the training data to improve lexical coverage during fine-tuning.
For evaluation, the same Quechua normalization is applied to both system
outputs and reference translations, ensuring that metric computation reflects
orthographic equivalence rather than spacing artifacts.

\paragraph{Aymara.}
For Aymara--Spanish (aym), we apply a deliberately conservative preprocessing
pipeline designed to improve data cleanliness while preserving surface
orthographic structure. This includes Unicode normalization (NFKC),
normalization of apostrophe variants, whitespace normalization, and removal
of empty or misaligned sentence pairs. To reduce obvious alignment noise,
we filter sentence pairs exhibiting extreme length mismatches using the same
length-ratio heuristic applied to Guarani, and remove exact duplicate pairs.

In addition, we address a systematic corpus artifact in which apostrophes—
used in common Aymara orthographies to mark ejective or glottalized consonants—
are separated from surrounding characters by spurious whitespace
(e.g., \texttt{jach 'a}, \texttt{t 'äw}, \texttt{qilqt 'am}). We apply a
deterministic orthographic normalization rule that merges intra-word
\texttt{letter~'~letter} patterns into a single token
(e.g., \texttt{jach'a}, \texttt{t'äw}), without introducing any additional
linguistic rewriting.

Despite improving overall data consistency, these preprocessing steps yield
limited gains in translation quality for Aymara. We hypothesize that this is
due to the language’s strongly agglutinative morphology, in which grammatical
information is encoded through productive suffix chains and phonemic contrasts.
Generic normalization combined with subword tokenization may fragment these
morphemes, motivating morpheme-aware preprocessing or segmentation strategies
as future work~\cite{mager2021findings}.

\subsection{Model Fine-tuning}
We fine-tune the multilingual mBART model (\texttt{facebook/mbart-large-50}) using the HuggingFace Transformers library. mBART is pretrained using a denoising autoencoding objective across 50 languages, enabling effective cross-lingual transfer in low-resource translation scenarios \cite{tang2020multilingual}. We use the \texttt{MBart50Tokenizer} with Spanish (\texttt{es\_XX}) as the source language and the appropriate target language tags (\texttt{aym\_XX} for Aymara and \texttt{gn\_XX} for Guarani).

To better support Guarani orthography, we extend the tokenizer vocabulary with Guarani-specific characters and frequent multi-character units, including nasal vowels and diacritics (\textit{\~{a}, \~{e}, \~{\i}, \~{o}, \~{u}}), the combining tilde, and common digraphs (\textit{ch, mb, ng}). After extending the tokenizer, the model’s embedding matrix is resized accordingly.

For Aymara--Spanish translation, we train using a learning rate of $2\times10^{-5}$, batch size 16, gradient accumulation of 4, and 20 epochs. For Guarani--Spanish translation, we use a learning rate of $3\times10^{-5}$, batch size 8, gradient accumulation of 4, and 15 epochs. All models are optimized using AdamW with a cosine learning rate schedule and warm-up. Early stopping with a patience of three evaluation intervals based on validation BLEU is applied to prevent overfitting. These configurations are consistent with recent AmericasNLP submissions \cite{gow2023sheffield, tonja2023enhancing}.

\subsection{Evaluation Metrics}
We evaluate translation quality primarily using chrF++, 
computed with the official AmericasNLP 2023 evaluation script.
While BLEU is monitored during training for early stopping, 
we do not report BLEU scores in our main results due to its 
known limitations for morphologically rich and agglutinative 
languages such as Guarani, Quechua, and Aymara\cite{popovic-2015-chrf}\cite{popovic-2017-chrf}.

\section{Experimental Results}

\subsection{Aymara--Spanish Translation}
For Aymara–Spanish translation, mBART fine-tuning yields limited improvements under both curated-only and synthetically augmented settings. While chrF++ increases modestly with synthetic data, overall performance remains low, consistent with prior findings for highly agglutinative languages under subword tokenization.
These results suggest that generic normalization and data augmentation are insufficient for Aymara, motivating morpheme-aware approaches discussed in Section 6.

\subsection{Guarani--Spanish Translation}
A similar trend is observed for Guarani--Spanish translation. The curated-only
model achieves a chrF++ score of 42.00 on the development set. Incorporating
synthetic data improves performance to 44.00 chrF++, yielding a gain of
+2.00 points. This result indicates that forward-translated synthetic data
provides consistent benefits even when training data is limited.

As with Aymara--Spanish, training dynamics for Guarani--Spanish exhibit
stable optimization behavior, with improved validation performance across
epochs when synthetic data is included. Together, these results demonstrate
that the benefits of synthetic augmentation extend beyond a single language
pair.

\subsection{Quechua--Spanish Translation}
For Spanish--Quechua translation, applying deterministic orthographic normalization yields substantial improvements in chrF++. In particular, resolving systematic intra-word spacing artifacts (e.g., \textit{ch aypiqa}, \textit{sin ch i}) reduces token fragmentation and improves character-level alignment. On the AmericasNLP 2023 development set, our model achieves a chrF++ score of 36.6, outperforming both the shared-task baseline and matching the best reported systems from the Sheffield submission. As in prior work, performance under n-gram-based metrics remains challenging due to Quechua’s agglutinative morphology, and we therefore emphasize chrF++ as the primary evaluation metric.

\subsection{Effect of Synthetic Data Augmentation}
Across both language pairs, synthetic data augmentation consistently improves chrF++ scores relative to curated-only training. These findings align with prior work showing that synthetic parallel data generated using high-capacity multilingual models can significantly enhance MT performance for indigenous and other low-resource languages. In line with results reported in recent AmericasNLP shared tasks, our experiments confirm that forward-translated synthetic data can serve as an effective and scalable strategy for improving NMT quality in data-scarce scenarios.

\subsection{Comparison with AmericasNLP 2023}
We compare our systems against the AmericasNLP 2023 baseline and the best per-language results reported by the University of Sheffield on the official development sets. Following prior work, we focus on chrF++ due to BLEU’s known limitations for agglutinative languages\cite{popovic-2015-chrf}\cite{popovic-2017-chrf}.

Top-performing submissions primarily relied on large multilingual models such as NLLB-200\cite{costa2022no}, often paired with back-translation or other forms of synthetic data generation \cite{gow2023sheffield, tonja2023enhancing}. In particular, the University of Sheffield system employed NLLB-200 with extensive back-translation and diverse parallel sources, achieving strong results across multiple language pairs \cite{gow2023sheffield}. These findings highlight the effectiveness of data-centric approaches in low-resource indigenous MT.

Our method aligns with this paradigm by leveraging synthetic parallel data generated via forward translation using NLLB-200, while fine-tuning a multilingual mBART model. Although our approach differs from prior work in its exclusive use of forward-translated synthetic data from the MultiScript30k pipeline, our results are consistent with trends observed in the shared task: well-curated synthetic data, when paired with robust multilingual models, can substantially improve translation quality for indigenous languages.

\section{Baseline Comparison}
\begin{table*}[t]
\centering
\small
\setlength{\tabcolsep}{10pt}
\begin{tabular}{lccc}
\hline
\textbf{Team / System (dev chrF)} & \textbf{aym} & \textbf{gn} & \textbf{quy} \\
\hline
2021 Baseline (Vázquez et al., 2021)        & 15.70 & 19.30 & 30.40 \\
2021 Best System                           & 28.30 & 33.60 & 34.30 \\
\hline
Andes                                     & 9.22  & --    & --    \\
CIC-NLP                                   & 19.05 & 21.75 & 35.62 \\
Helsinki-NLP                              & 33.44 & 40.42 & 37.19 \\
LCT-EHU                                   & --    & --    & 38.59 \\
LTLAmsterdam                              & 25.23 & 32.89 & 36.81 \\
PlayGround                                & 29.98 & 33.17 & 34.38 \\
Sheffield                                 & 36.24 & 39.34 & 39.52 \\
\hline
$\uparrow$ 2021 (Best -- Baseline)         & +12.60 & +14.30 & +3.90 \\
$\uparrow$ 2023 (Best -- 2021 Best)        & +7.94  & +6.82  & +5.22 \\
\hline
Ours (Curated-only)                        & 26.85 & 42.00 & 37.12 \\
Ours (Curated + Synthetic)                 & 30.82 & 44.00 & 37.83 \\
\hline
\end{tabular}
\caption{Development-set chrF++ comparison for Aymara (aym), 
Guarani (gn), and Quechua (quy) on the AmericasNLP benchmarks\cite{ebrahimi-etal-2023-findings}. 
Prior results are taken from the official shared-task reports.}

\label{tab:baseline_comp}
\end{table*}

To contextualize our results, we compare our system against the best-performing submission reported in the AmericasNLP 2023 shared task for Guarani--Spanish translation \cite{ebrahimi2023findings}. While differences in training data and augmentation strategies prevent a strictly controlled comparison, this provides a meaningful reference point within the established evaluation framework. Our results indicate that synthetic data augmentation improves performance over a curated-only baseline and yields competitive chrF++ scores relative to reported shared-task systems.Table~\ref{tab:baseline_comp} compares our dev-set chrF++ scores with the AmericasNLP 2023 baseline and the best per-language results reported by the University of Sheffield submission. Table~3 shows that synthetic data augmentation improves chrF++ across all
three languages. Gains are largest for Aymara (+3.97 chrF++), followed by
Guarani (+2.00), while Quechua shows smaller but consistent improvements
(+0.71), reflecting its already large curated training set. Although Aymara shows the largest relative chrF++ gain, absolute performance
remains low, reinforcing the need for morphology-aware modeling.
 While direct comparison is limited by differences in model architecture and training configurations, this provides context for the effectiveness of our synthetic data augmentation approach.

\section{Discussion}

Our results demonstrate that fine-tuning a multilingual mBART model with synthetic data augmentation is an effective strategy for improving MT performance for low-resource indigenous languages such as Aymara and Guarani. The observed gains in chrF++ indicate that forward-translated synthetic data can meaningfully complement small curated corpora, supporting better generalization in data-scarce settings. Language-specific preprocessing further contributes to these improvements by reducing noise while preserving important morphological and phonological structure.

The effectiveness of our approach is consistent with findings from recent AmericasNLP shared tasks, where multilingual pre-trained models combined with synthetic data generation have emerged as strong baselines for indigenous language MT. While some systems—such as the University of Sheffield submission—achieved strong performance using larger NLLB-200 models and back-translation \cite{gow2023sheffield}, our results suggest that competitive improvements can also be obtained using mBART paired with forward-translated synthetic data. This highlights the flexibility of data-centric augmentation strategies across different model architectures.

Despite these encouraging results, several limitations remain. Our evaluation relies exclusively on automatic metrics, which may not fully capture cultural nuance or idiomatic correctness in indigenous languages \cite{mager2023ethical}. Future work should therefore incorporate human evaluation involving proficient speakers. Additionally, further investigation into synthetic data quality, diversity, and filtering strategies—as well as alternative augmentation methods such as back-translation—may yield additional gains. Finally, incorporating explicit linguistic knowledge of Aymara and Guarani into model design or preprocessing pipelines \cite{garvin1972urbanization} represents a promising direction for improving translation accuracy while ensuring ethical and community-centered deployment of MT systems \cite{mager2023ethical}.

\section{Conclusion and Future Work}
In this work, we investigated synthetic data augmentation and language-specific preprocessing for low-resource indigenous language machine translation, focusing on Guarani–Spanish and Quechua–Spanish, with diagnostic experiments on Aymara.

Future work will extend this framework in several directions. First, we plan to conduct human evaluations to better assess translation quality and cultural adequacy. Second, we aim to explore additional indigenous languages, including Aymara, with more targeted preprocessing and data augmentation strategies.
A promising direction for future work is the incorporation of visual context for indigenous language translation. Resources such as MultiScript30k provide aligned image–caption–translation triples for languages including Guarani, enabling investigation of multimodal machine translation in low-resource settings. Future experiments will explore whether visual grounding can improve translation robustness, reduce ambiguity, or accelerate model convergence when parallel text is scarce. This direction aligns with recent interest in data-efficient and multimodal approaches for low-resource MT.

\section{Limitations}
This study has several limitations. First, evaluation relies exclusively on automatic metrics (chrF++), which may not fully capture translation adequacy, fluency, or culturally grounded meaning in indigenous languages. Human evaluation by proficient speakers is therefore necessary to assess real-world translation quality. While orthographic normalization improves Quechua translation, it relies on deterministic rules and may not generalize to dialectal variation or other Quechua varieties. Second, synthetic data is generated automatically and may introduce systematic biases or translation artifacts, despite filtering and preprocessing. Finally, while we report detailed experiments for Guarani--Spanish, results for other indigenous languages such as Aymara remain limited and are not fully explored in this work.

%\section{Results}
% Cosine Similarity paragraph

% \section{Future Work}

% \section{Acknowledgments}

\bibliography{citations}
\bibliographystyle{acl_natbib}

\end{document}